\documentclass[a4paper,fleqn]{cas-sc}

\usepackage[authoryear,longnamesfirst]{natbib}

\usepackage{multirow}
\usepackage{algorithm}
\usepackage{algpseudocode}
\usepackage{newfloat}
\usepackage{listings}

\algnewcommand{\COMMENT}[1]{\hfill \textit{// #1}}
\makeatletter
\algnewcommand{\NoNumber}{\renewcommand{\theALG@line}{}}
\algnewcommand{\PhaseComment}[1]{\Statex \NoNumber\textbf{#1}}
\algnewcommand{\StepComment}[1]{\Statex \NoNumber\hspace{\algorithmicindent}\textit{#1}}
\makeatother

\def\tsc#1{\csdef{#1}{\textsc{\lowercase{#1}}\xspace}}
\tsc{WGM}
\tsc{QE}
\tsc{EP}
\tsc{PMS}
\tsc{BEC}
\tsc{DE}

\newdefinition{definition}{Definition}
\newdefinition{remark}{Remark}
\newproof{proof}{Proof}

\begin{document}
\let\WriteBookmarks\relax
\def\floatpagepagefraction{1}
\def\textpagefraction{.001}
\shorttitle{Kinematic-Space Completion for Expression-Controlled 3D Gaussian Avatar Animation}
\shortauthors{H. Wei et~al.}

\title [mode = title]{LiftAvatar: Kinematic-Space Completion for Expression-Controlled 3D Gaussian Avatar Animation}                      



\author[1]{Hualiang Wei}[type=editor,
                        orcid=0009-0001-7417-2687]

\ead{weihl23@mails.jlu.edu.cn}

\credit{Writing – review \& editing, Writing – original draft, Methodology, Software, Conceptualization, Data curation, Formal analysis, Visualization, Validation}

\affiliation[1]{organization={College of Computer Science and Technology, Jilin University},
                addressline={No. 2699 Qianjin Street}, 
                city={Changchun},
                postcode={130012}, 
                state={Jilin},
                country={China}}

\author[2]{Shunran Jia}[
                        orcid=0009-0006-1041-3484]

\ead{shunran@socialbook.io}

\credit{Formal analysis, Resources, Software}

\affiliation[2]{organization={Impressed Inc DBA SocialBook},
                addressline={950 Tower Ln},
                city={Foster City},
                postcode={94404},
                state={California},
                country={USA}}

\author[3]{Jialun Liu}[
                        orcid=0009-0001-6161-3842]

\ead{liujialun95@gmail.com}
\credit{Writing – review \& editing, Investigation}

\affiliation[3]{organization={Institute of Artificial Intelligence of China Telecom},
            addressline={Zhonghui Building, Dongcheng District}, 
            postcode={100007}, 
            state={Beijing},
            country={China}}

\author[1]{Wenhui Li}[
                        orcid=0000-0001-6490-9852]
                        
\cormark[1]
\ead{liwh@jlu.edu.cn}
\credit{Writing – review \& editing, Supervision, Funding acquisition}

\cortext[cor1]{Corresponding author}

\begin{abstract}
We present LiftAvatar, a new paradigm that completes sparse monocular observations in kinematic space (e.g., facial expressions and head pose) and uses the completed signals to drive high-fidelity avatar animation. LiftAvatar is a fine-grained, expression-controllable large-scale video diffusion Transformer that synthesizes high-quality, temporally coherent expression sequences conditioned on single or multiple reference images. The key idea is to lift incomplete input data into a richer kinematic representation, thereby strengthening both reconstruction and animation in downstream 3D avatar pipelines. To this end, we introduce (i) a multi-granularity expression control scheme that combines shading maps with expression coefficients for precise and stable driving, and (ii) a multi-reference conditioning mechanism that aggregates complementary cues from multiple frames, enabling strong 3D consistency and controllability. As a plug-and-play enhancer, LiftAvatar directly addresses the limited expressiveness and reconstruction artifacts of 3D Gaussian Splatting-based avatars caused by sparse kinematic cues in everyday monocular videos. By expanding incomplete observations into diverse pose-expression variations, LiftAvatar also enables effective prior distillation from large-scale video generative models into 3D pipelines, leading to substantial gains. Extensive experiments show that LiftAvatar consistently boosts animation quality and quantitative metrics of state-of-the-art 3D avatar methods, especially under extreme, unseen expressions.
\end{abstract}




\maketitle

\section{Introduction}

The rapid progress of virtual reality, augmented reality, and telepresence is fueling an increasing demand for realistic, animatable, and real-time renderable 3D digital human head models~\cite{cheng2014children,healey2021mixed,kachach2020virtual,li2021vmirror}. Commonly referred to as \emph{3D head avatars}, these models seek to reconstruct high-fidelity geometry and fine-grained appearance from monocular or multi-view observations, enabling controllable synthesis across viewpoints, head poses, and facial expressions. Such capabilities are central to immersive communication~\cite{li2020volumetric,ma2021pixel,tran2024voodoo}, entertainment~\cite{sklyarova2023haar,zhu2020reconstructing}, and digital content creation~\cite{li20233d,naruniec2020high}.

Recent years have pushed 3D head avatar reconstruction along the quality-efficiency Pareto frontier. In particular, 3D Gaussian Splatting (3DGS)~\cite{DBLP:journals/tog/KerblKLD23} has emerged as a strong representation, offering high-quality rendering with real-time performance, and has been widely adopted in avatar reconstruction pipelines~\cite{DBLP:journals/corr/abs-2404-01053,DBLP:conf/cvpr/ShaoWLWL00024,DBLP:conf/siggraph/Chen0LXZYL24,DBLP:conf/cvpr/XuCL00ZL24}. Despite this success, 3DGS-based avatars remain \emph{data-hungry}: the reconstruction quality and downstream animation fidelity strongly depend on whether the training observations sufficiently cover the underlying \emph{state space} of the subject. This limitation is especially pronounced for monocular videos captured in everyday settings, where head pose changes and facial expressions are often limited in diversity. We refer to these motion-related cues, including expression dynamics and head pose, as \emph{kinematic information}. When kinematic coverage is sparse, 3DGS models tend to overfit to the observed motion patterns, leading to artifacts in reconstruction and brittle animation that collapses under \emph{unseen} or \emph{extreme} expressions.

\begin{figure}
    \centering
    \includegraphics[width=0.95\linewidth]{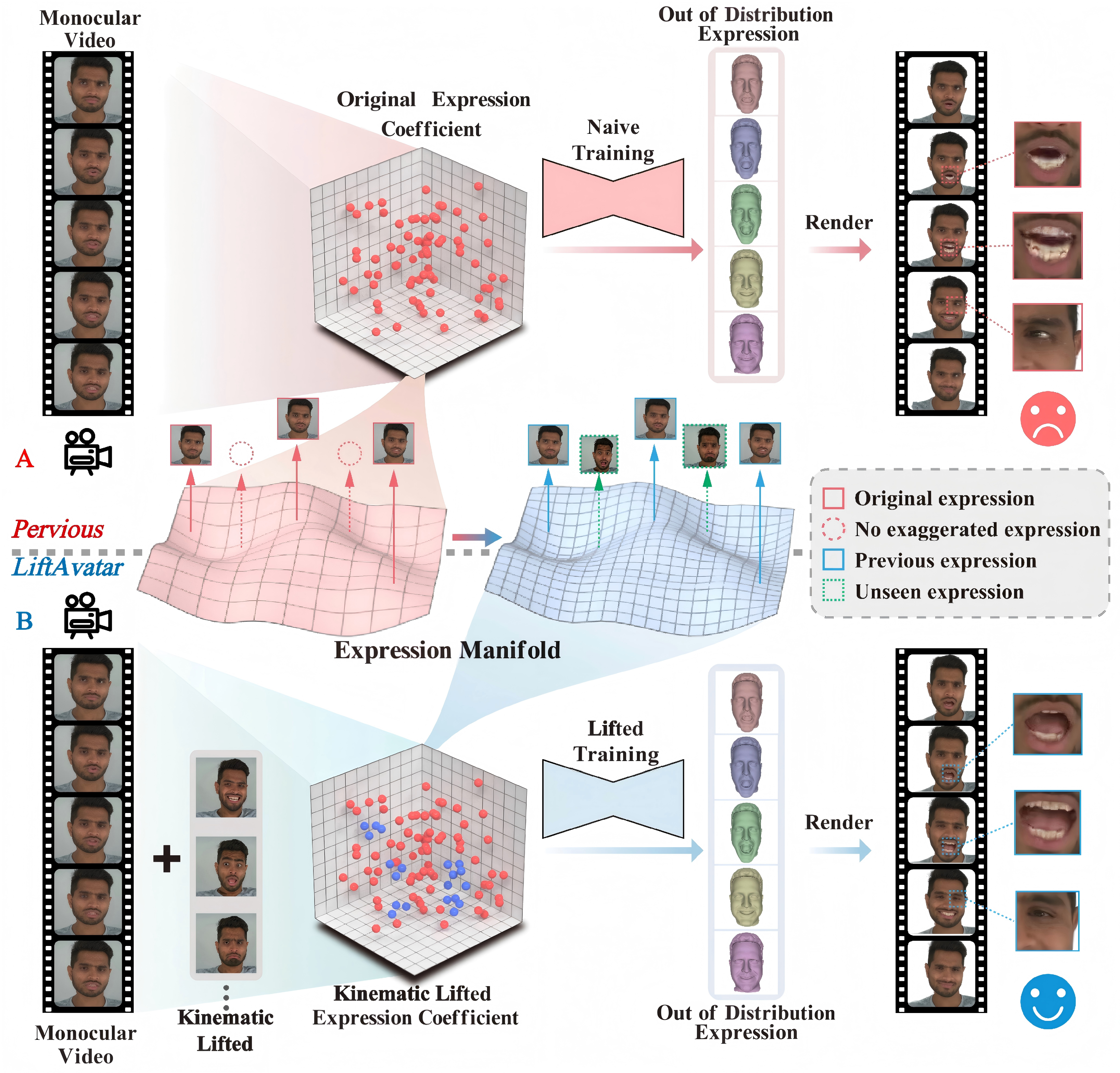}     
    \caption{We propose a novel kinematic lifted framework, LiftAvatar, to complement the facial expressions and head poses of the monotonous input video. LiftAvatar can promote subsequent reconstruction and driving tasks, resulting in significant improvements.}
    \label{Figure_1}
\end{figure}

Most existing efforts address related but different bottlenecks. One line of work improves \emph{low-level capture quality}, such as motion blur removal or robust camera pose estimation from handheld videos~\cite{DBLP:conf/cvpr/MaL0ZWWS22,DBLP:conf/eccv/ZhaoWL24,DBLP:conf/iccv/LiuWHTZV23,DBLP:conf/cvpr/Wu0TR0SFGL25}. Another line focuses on \emph{sparse-view} or \emph{few-shot} reconstruction, including single-image 3D head avatar creation~\cite{deng2024portrait4d,deng2024portrait4d1,li2023generalizable,tran2024voodoo1,bao2024geneavatar}. These advances demonstrate that plausible geometry and appearance can be recovered from limited viewpoints. However, even when the video is stable and contains many frames, a fundamental issue often remains: the observations may still be \emph{kinematically incomplete}. Typical monocular capture sequences contain only a small subset of expressions (e.g., neutral or mild smile) and limited pose range, leaving large regions of the expression-pose space unobserved. This \emph{kinematic-space incompleteness} is not directly solved by deblurring, pose refinement, or few-view geometry priors, and it becomes a primary failure mode for 3DGS-based avatar animation: the reconstructed avatar appears stiff and produces unrealistic deformations when driven beyond the narrow motion manifold seen during training.

We introduce \textbf{LiftAvatar}, a plug-and-play kinematic-space completion framework that \emph{lifts} an expression-sparse monocular video into a richer set of pose and expression observations using a large-scale video diffusion transformer. The key idea is to leverage generative priors to \emph{expand} the training observations along the kinematic dimensions while preserving the subject identity and appearance details, thereby providing the downstream 3D reconstruction pipeline with sufficiently diverse and semantically meaningful motion coverage. In contrast to prior avatar video generation models that often suffer from temporal flickering and coarse motion control~\cite{DBLP:conf/cvpr/XuYZZJHJZCTL0L25,DBLP:journals/corr/abs-2412-00733,DBLP:journals/corr/abs-2407-03168,DBLP:journals/corr/abs-2402-17485,DBLP:conf/nips/SiarohinLT0S19}, LiftAvatar is designed specifically for \emph{precision} and \emph{consistency} required by 3D training.

LiftAvatar is enabled by two core designs. First, for fine-grained and anatomically meaningful control, we adopt the Neural Parametric Head Model (NPHM)~\cite{DBLP:conf/cvpr/GiebenhainKGRAN23} rather than linear parametric models such as FLAME~\cite{DBLP:journals/tog/LiBBL017} or 3DMM~\cite{DBLP:conf/siggraph/BlanzV99}. We propose a \emph{multi-granularity kinematic conditioning} scheme that jointly injects NPHM shading maps and expression coefficient vectors into the diffusion transformer, enabling precise control over head pose and facial dynamics while retaining photorealistic micro-details (e.g., wrinkles and lip-corner shapes). Second, to maximize identity and appearance grounding, LiftAvatar supports \emph{multi-reference conditioning} with an arbitrary number of input frames. We integrate reference cues via patch embedding expansion and CLIP~\cite{radford2021learning} cross-attention, allowing the model to aggregate complementary information across frames (identity, texture, hair, illumination) and reducing over-reliance on priors that can cause subject drift. To further strengthen temporal coherence, we adopt the flow-matching training objective from Wan~\cite{DBLP:journals/corr/abs-2503-20314}. Importantly, LiftAvatar acts as a "rocket booster" during training: once the 3D avatar is learned, LiftAvatar can be removed, incurring no additional cost at inference time for the final 3D system.

In summary, our contributions are:
\begin{itemize}
    \item We propose {LiftAvatar}, a kinematic-space completion framework built on a large-scale video diffusion transformer that enriches expression-sparse monocular videos with diverse, high-fidelity head poses and fine-grained expressions, directly addressing the key bottleneck in monocular 3D avatar animation.
    \item We introduce a {multi-granularity expression control} scheme that leverages NPHM shading maps and expression coefficients to achieve high-precision, detail-preserving facial motion control suitable for downstream 3D training.
    \item We design a {multi-reference conditioning} mechanism that fuses an arbitrary number of reference frames via patch embedding expansion and CLIP cross-attention, improving identity grounding and reducing artifacts.
    \item Extensive experiments show that LiftAvatar substantially improves both visual quality and quantitative metrics of state-of-the-art 3D avatar pipelines, MonoGaussianAvatar, especially under extreme and unseen expressions, while preserving the inference efficiency of the final 3D avatar model.
\end{itemize}

\section{Related Work}

\subsection{Data Augmentation for 3D Reconstruction}
Data augmentation is widely used to improve robustness and visual quality in 3D reconstruction, especially for NeRF~\cite{DBLP:journals/cacm/MildenhallSTBRN22} and 3D Gaussian Splatting (3DGS)~\cite{DBLP:journals/tog/KerblKLD23}. Existing methods can be roughly divided into three directions.
Low-level enhancement methods primarily address degradations such as noise, motion blur, and unstable camera motion, aiming to improve the quality of observed frames for more stable reconstruction~\cite{DBLP:conf/cvpr/MaL0ZWWS22,DBLP:conf/eccv/LeeLSAP24}. While effective when the input is corrupted, these methods are limited to restoring what is already observed, and thus cannot resolve structural incompleteness, such as missing viewpoints or sparse expression coverage.
A second line of work distills pretrained generative priors into 3D representations. Score distillation sampling (SDS) leverages diffusion models to hallucinate missing observations and enables reconstruction from sparse inputs~\cite{DBLP:journals/corr/abs-2310-15110,DBLP:conf/iclr/PooleJBM23,DBLP:conf/iccv/LiuWHTZV23}. Despite strong results, SDS-based optimization is often unstable and can lead to over-saturation, over-smoothing, and blurred details.
A third direction emphasizes efficiency via feed-forward 3D inference, bypassing per-scene optimization by directly predicting 3D attributes from images. Representative methods such as DUSt3R~\cite{wang2024dust3r} and VGGT~\cite{wang2025vggt} achieve fast inference, but in head avatar settings they typically struggle with fine-grained dynamic motion modeling and are often constrained to fixed-form inputs. This limits their ability to exploit variable-length real-world observations such as monocular videos or sparse multi-view sets, leading to underutilization of available data.

In contrast to these paradigms, our goal is to address kinematic incompleteness, a common but under-explored bottleneck in monocular head avatar reconstruction. Rather than restoring low-level degradations or hallucinating missing viewpoints, we explicitly complete the pose and expression space by synthesizing semantically meaningful, identity-consistent kinematic variations. This enriches the training distribution for downstream 3D reconstruction and improves both reconstruction robustness and animation fidelity.

\subsection{Head Avatar Reconstruction and Animation}
The pursuit of high-fidelity, animatable 3D head avatars has progressed from parametric face models to neural rendering and generative priors. Early systems relied on 3DMM~\cite{DBLP:conf/siggraph/BlanzV99} and FLAME~\cite{DBLP:journals/tog/LiBBL017} with learning-based refinements for expression and pose control~\cite{DBLP:conf/avss/PaysanKARV09,cudeiro2019capture,DBLP:conf/cvpr/Yang0WHSYC20,feng2021learning,DBLP:conf/cvpr/WangCYMLL22,danvevcek2022emoca,ma2024cvthead}, but their linear parameterization limits fine-scale detail. NeRF-based avatars~\cite{DBLP:journals/cacm/MildenhallSTBRN22,athar2022rignerf,guo2021ad,liu2022semantic} significantly improved photorealism, yet remain optimization-heavy and slow for real-time use. More recently, 3DGS~\cite{DBLP:journals/tog/KerblKLD23} enabled high-quality real-time rendering and has been widely adopted for controllable head avatars~\cite{DBLP:conf/siggraph/Chen0LXZYL24,DBLP:conf/cvpr/XuCL00ZL24,qian2024gaussianavatars,wang2025gaussianhead}; however, these pipelines are data-hungry and often fail to generalize when monocular training videos exhibit limited pose or expression diversity.
Generative priors have also been exploited to compensate for incomplete observations. GAN-based approaches such as EG3D~\cite{DBLP:conf/cvpr/ChanLCNPMGGTKKW22} and ray conditioning~\cite{DBLP:conf/iccv/ChenHYZD23} leverage 2D priors~\cite{DBLP:conf/cvpr/KarrasLA19} for novel-view synthesis, while diffusion-based methods further extend to view and expression generation~\cite{DBLP:conf/cvpr/XuYZZJHJZCTL0L25,DBLP:conf/cvpr/ChenMWP024,DBLP:conf/cvpr/KirschsteinGN24,DBLP:journals/corr/abs-2409-16990,DBLP:journals/corr/abs-2409-18083}. Related controllable generation frameworks model motion priors for long-horizon facial dynamics~\cite{shenlong} and unify conditional generation for pose-guided humans~\cite{shen2024imagpose} and customizable virtual dressing~\cite{shen2025imagdressing}. Nonetheless, generating unseen expressions or extreme motions from limited 2D evidence remains challenging, often causing temporal inconsistency, texture distortion, or identity drift.

To improve efficiency, feed-forward avatar models regress animatable heads from one or a few images within seconds~\cite{he2025lam,kirschstein2025avat3r}, but are commonly constrained by fixed-form inputs and cannot fully exploit variable-length real-world videos. Overall, existing paradigms work best when the input already contains rich kinematic variation. Our method targets this bottleneck by combining a high-fidelity facial parameterization, NPHM~\cite{DBLP:conf/cvpr/GiebenhainKGRAN23}, with a large-scale video diffusion transformer to synthesize identity-consistent pose and expression variations, completing the kinematic space as a plug-and-play pre-processor for downstream 3D avatar pipelines.

\begin{figure}
    \centering
    \includegraphics[width=0.95\linewidth]{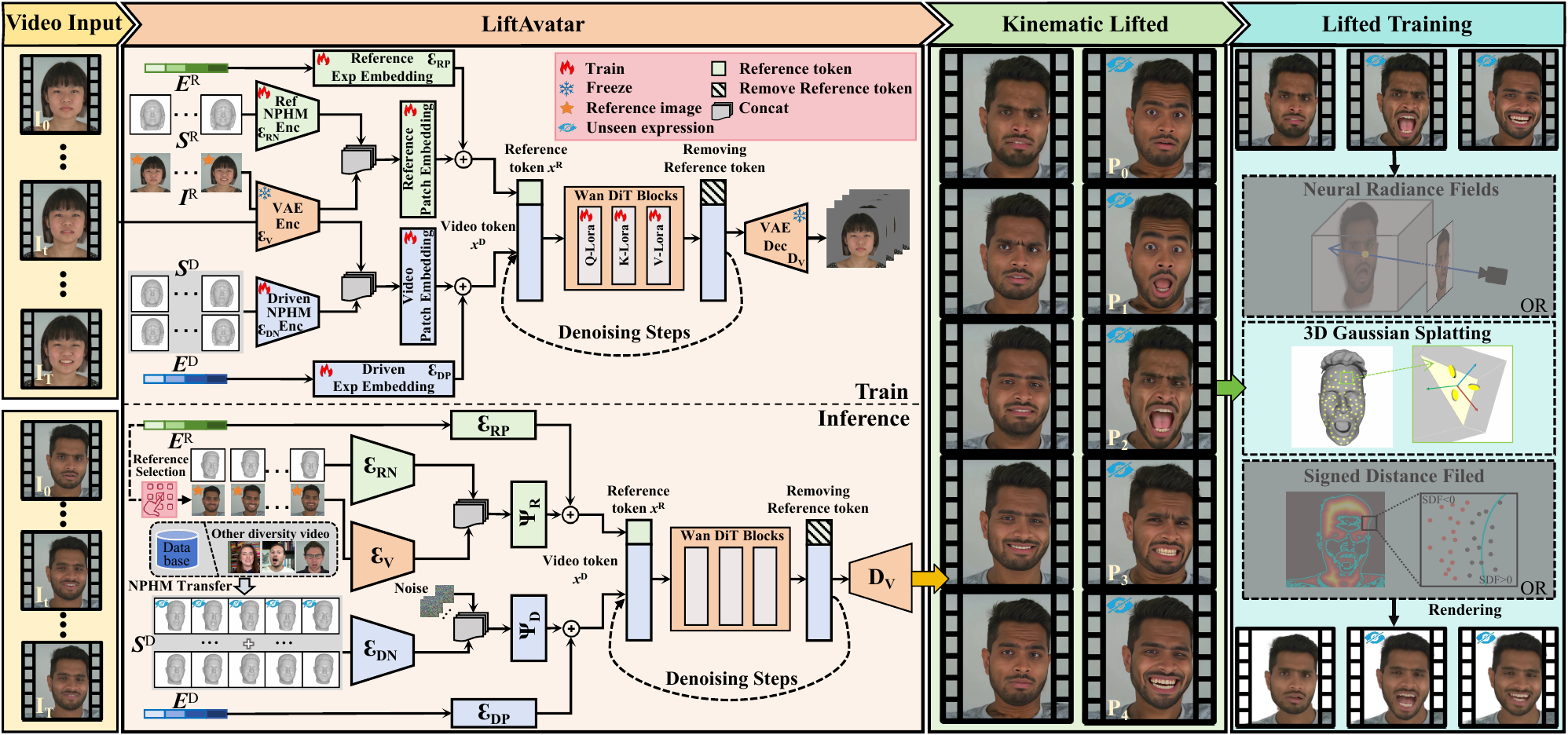}
    \centering
        \caption{Method Pipeline. LiftAvatar is a high-precision, expression‑controlled video diffusion transformer that enriches sparse observations to boost downstream avatar performance. It conditions on three groups of inputs: reference information $(I^R, S^R, E^R)$ from the input video; driving information $(S^D, E^D)$ for the target motion; and the ground‑truth driving video $V^D$ during training (replaced by noise at inference). The reference images are encoded by a pre‑trained VAE, and the Reference NPHM Encoder encodes their shading maps $S^R$. These features are concatenated, projected via Reference Patch Embedding, and summed with the embedded expression coefficients $E^R$ to form the reference token $x^R$. Likewise, the Driven NPHM Encoder processes the driving shading maps $S^D$; its output is projected by Driven Patch Embedding and combined with the embedded $E^D$ to produce the driving token $x^D$. The tokens $x^R$ and $x^D$ are concatenated and fed as a unified condition into the Wan2.1~\cite{DBLP:journals/corr/abs-2503-20314} video diffusion transformer backbone. Optimized with a flow‑matching objective, the model synthesizes high‑fidelity, temporally coherent videos that accurately follow the driving signals, thereby completing the kinematic space of the original input.}
    \label{Figure_2}
\end{figure}

\section{Methodology}

\subsection{Overview}

Given a daily video $V = \{ i_0, i_1, \cdots, i_T\}$, where $T$ represents the total number of frames (typically featuring neutral facial expressions and minimal head movement), the 3D avatar reconstructed from such a video usually lacks expressiveness. In particular, when driven by unseen expressions, it often suffers from severe artifacts such as geometric collapse. However, this type of monotonous video is prevalent both on the Internet and in personal recordings. Our goal is to address this challenge by employing kinematic space completion techniques. To this end, we introduce \textbf{LiftAvatar}, a novel kinematic lifted training framework. It employs a powerful large-scale video diffusion transformer to lift the facial expressions and head poses from limited input, thereby significantly expanding the capacity of the input information. Using these lifted kinematic priors, LiftAvatar substantially enhances the reconstruction and animation quality of downstream 3D avatars. Crucially, the framework is decoupled from any specific head avatar reconstruction method, serving as a plug-and-play module compatible with a wide range of existing 3D avatar techniques.

\subsection{Framework of LiftAvatar}
\label{sec:augment}

To achieve an effective increase in training data, the model must have sufficient fidelity and generation quality, as lower quality kinematic lifted data compared to existing data would adversely affect 3D avatar training. Most prior methods adopt ReferenceNet architectures~\cite{DBLP:journals/corr/abs-2402-17485}, but experimental observations indicate that such frameworks suffer from suboptimal temporal consistency. Therefore, LiftAvatar employs the state-of-the-art open source video model (i.e., Wan2.1~\cite{DBLP:journals/corr/abs-2503-20314}) as its foundational framework. While Wan2.1~\cite{DBLP:journals/corr/abs-2503-20314} delivers compelling temporal consistency, its control precision falls short. To address this, we develop a model based on the Wan2.1~\cite{DBLP:journals/corr/abs-2503-20314} framework, as depicted in Figure~\ref{Figure_2}. Our model generates high-precision and high-fidelity videos by employing a multi-granularity expression control scheme and a multi-reference conditioning mechanism. LiftAvatar accepts six types of inputs: reference images $I^R = \{i^R_0, ..., i^R_M\}$, the corresponding NPHM~\cite{DBLP:conf/cvpr/GiebenhainKGRAN23} shading map of reference images $S^R = \{s^R_0, ..., s^R_M\}$, and NPHM expression parameters $E^R=\{e^R_0,...,e^R_M\}$. A sequence of driving NPHM shading maps $S^D=\{s^D_0,...,s^D_N\}$; and the corresponding NPHM~\cite{DBLP:conf/cvpr/GiebenhainKGRAN23} expression coefficients $E^D=\{e^D_0,...,e^D_N\}$. During training, a target driving video $V^D=\{I^D_0,...,I^D_N\}$ is additionally provided as input, which is replaced by noise $n^D$ during the inference phase. $N$ denotes the number of reference images and $M$ denotes the length of the driving sequence. LiftAvatar accurately drives the reference image into a corresponding video sequence guided by the driving expression sequence. It should be noted that during training, the driving signal $S^D$ aligns with $V^D$, which is extracted from the input video $V$. However, during inference, to enrich the expressions in $V$, $S^D$ and $E^D$ are transferred from other expression-rich videos.



\noindent\textbf{Fine-grained Expression Control.}
Unlike mainstream avatar models that rely primarily on linear parametric models (e.g., FLAME~\cite{DBLP:journals/tog/LiBBL017} or 3DMM~\cite{DBLP:conf/siggraph/BlanzV99}), this paper adopts NPHM~\cite{DBLP:conf/cvpr/GiebenhainKGRAN23} for high-precision expression representation. To this end, we design a multi-granularity expression control scheme. We first use shading maps as fine-grained guidance signals, which effectively capture detailed facial variations. However, relying solely on shading maps proves inadequate for achieving the desired level of control precision. Therefore, we incorporate NPHM~\cite{DBLP:conf/cvpr/GiebenhainKGRAN23} expression coefficients as a complementary control mechanism to provide more precise and structured guidance.

To achieve efficient and low-computational injection of the two types of control information mentioned above, we designed two lightweight modules: the NPHM Encoder and the NPHM Exp Embedding. The NPHM Encoder is divided into the Reference NPHM Encoder $\mathcal{E}_{RN}$ and the Driven NPHM Encoder $\mathcal{E}_{DN}$. The former encodes shading maps $S^R$ corresponding to reference images, while the latter encodes shading maps $S^D$ corresponding to driven sequences. Both are small encoders composed of 3D convolutions, which encode shading maps into latent codes aligned with the latent DiT. Given the different nature of our inputs, it has two instantiations:

\textit{(1) Driven NPHM Encoder.} The structure of the Driven NPHM Encoder consists of 7 Conv3D layers, which transform the input shading map of dimensions $B\times3\times F\times512\times512$ into a latent code of dimensions $B\times16\times \frac{F}{4}\times64\times64$, where $B$ denotes the batch size and $F$ the number of input frames. Its core architecture employs multi-level downsampling blocks, progressively compressing spatial dimensions and expanding the number of channels through stacked Conv3D layers, ultimately outputting condition embedding features aligned with video DiT of Wan2.1~\cite{DBLP:journals/corr/abs-2503-20314}. Since the driven NPHM~\cite{DBLP:conf/cvpr/GiebenhainKGRAN23} constitutes a continuous video sequence, temporal information extraction is required. Thus, we designed the Driven NPHM Encoder as a compact network composed of Conv3D layers. SiLU activation functions are employed to enhance non-linearity. The output layer generates latent code using Conv3D, matching the input dimensions of video DiT.

\textit{(2) Reference NPHM Encoder.} In contrast to the driven NPHM~\cite{DBLP:conf/cvpr/GiebenhainKGRAN23} sequence, the NPHM~\cite{DBLP:conf/cvpr/GiebenhainKGRAN23} shading maps of the reference images do not constitute a temporally relevant sequence. Consequently, temporal modeling is unnecessary. Thus, the Reference NPHM Encoder is designed as a straightforward six-layer Conv2D convolutional neural network, which transform the input shading map of dimensions $B\times3\times512\times512$ into a latent code of dimensions $B\times16\times64\times64$, where $B$ denotes batch size. This encoder incorporates three downsampling stages, each implemented by Conv2D layers with $stride=2$.

The NPHM Exp Embeddings are also divided into two parts: one (i.e., $\mathcal{E}_{RP}$) for reference images and one (i.e., $\mathcal{E}_{DP}$) for driven sequences. Both consist of a single MLP layer that aligns NPHM~\cite{DBLP:conf/cvpr/GiebenhainKGRAN23} expression coefficients with the features after patch embedding in the Wan2.1~\cite{DBLP:journals/corr/abs-2503-20314} model. $S^R$ is encoded by $\mathcal{E}_{RN}$ into latent code and concatenated with the latent code of $I^R$ processed by Wan2.1's~\cite{DBLP:journals/corr/abs-2503-20314} VAE Encoder $\mathcal{E}_{V}$ along the channel dimension to obtain the feature representation of the reference information. This feature is then embedded into the latent space of Wan2.1~\cite{DBLP:journals/corr/abs-2503-20314} through the Reference Patch Embedding layer $\Psi_{R}$. Subsequently, it is summed with the $\mathcal{E}_{RP}$ encoded NPHM~\cite{DBLP:conf/cvpr/GiebenhainKGRAN23} expression parameters to generate the reference token $x^R$. $x^R$ simultaneously encodes fine-grained control information and coarse-grained expression information, improving the precision of the reference input. Driving information undergoes a similar process: it is encoded and extracted via $\mathcal{E}_{DN}$, $\mathcal{E}_{DP}$ and $\Psi_{D}$ to obtain the driving token $x^D$. $x^R$ and $x^D$ are concatenated along the dimension axis and fed into the Wan2.1~\cite{DBLP:journals/corr/abs-2503-20314} to achieve controlled generation of the driven video.

\noindent\textbf{High-fidelity Generation.}
Unlike previous approaches~\cite{DBLP:journals/corr/abs-2407-03168, DBLP:conf/cvpr/XuYZZJHJZCTL0L25} that predominantly rely on a single reference image, which fails to convey complex expressions and poses simultaneously, thereby forcing heavy dependence on prior knowledge or resulting in over-smoothed details, this paper proposes LiftAvatar which supports the flexible injection of arbitrary reference images. To this end, we design a multi-reference conditioning mechanism. By leveraging multiple reference inputs, our method provides a richer and more precise source of identity and appearance details. This significantly reduces the reliance on learned priors for \textquotedbl{}borrowing\textquotedbl{} details and effectively mitigates the tendency to generate over-smoothed outputs, thereby aligning with our objective of high-fidelity generation.

\textit{(1) Reference Image Selection.} Selecting informative reference frames from in-the-wild videos is critical for final quality. Instead of manual or random selection, we propose an automatic strategy based on K-means clustering to maximize the expression diversity and informational coverage of the reference set. Specifically, we extract the NPHM~\cite{DBLP:conf/cvpr/GiebenhainKGRAN23} expression coefficients (a 100-dimensional vector) for each frame of the input video $V$ and cluster them into $k$ groups based on expression similarity. For each resulting cluster, the frame whose coefficient vector is closest to the cluster centroid is selected as a reference image. This approach ensures that the selected references collectively represent the broadest spectrum of expressions present in $V$, minimizing redundancy and reducing the model's need to \textquotedbl{}guess\textquotedbl{} missing details.

Specifically, the K-means algorithm operates as follows. We first load the expression coefficients for all frames and randomly initialize $k$ cluster centers $C=\{\mathbf{e}_1,\mathbf{e}_2,\ldots,\mathbf{e}_k\}$. Then, for each sample $\mathbf{e}_i$, we compute its Euclidean distance to every cluster center and assign it to the nearest one:
\begin{equation}
\text{label}_i = \arg\min_k \|\mathbf{e}_i - \mathbf{e}_k\|^2.
\end{equation}
After assignment, the cluster centers are updated by taking the mean of all samples within each cluster. Then, for each cluster, the frame with the smallest distance to the cluster center is selected as a reference:
\begin{equation}
\mathbf{e}_k = \frac{1}{N_k} \sum_{i \in \text{cluster}_k} \mathbf{e}_i,
\end{equation}
where $N_k$ denotes the number of samples in the $k$-th cluster.

\textit{(2) Multi-Reference Injection.} To effectively inject the selected reference images into the generation process, we employ two complementary approaches:
1) Latent Code Injection: The reference image is encoded by Wan2.1’s~\cite{DBLP:journals/corr/abs-2503-20314} VAE encoder, and the resulting latent code is injected into the input of the Wan2.1~\cite{DBLP:journals/corr/abs-2503-20314} model. However, this method suffers from catastrophic forgetting.
2) Extended CLIP~\cite{radford2021learning} Context: We extend the original CLIP~\cite{radford2021learning} context to a multi-frame CLIP~\cite{radford2021learning} context, thereby generalizing the single-image injection mechanism to support an arbitrary number of images.
Through the combination of these two injection strategies, where latent codes provide robust identity grounding and the extended CLIP context preserves precise appearance details, our model achieves high-fidelity generation of the driven video.

\noindent\textbf{Training Objective.}
LiftAvatar builds upon the Wan2.1~\cite{DBLP:journals/corr/abs-2503-20314} video generation framework by fine-tuning all LoRA modules in the attention layers of the pre-trained Wan~\cite{DBLP:journals/corr/abs-2503-20314} model. The following components are trained from scratch: $\mathcal{E}{RN}$, $\mathcal{E}{RP}$, $\mathcal{E}{DN}$, $\mathcal{E}{DP}$, $\Psi_R$, and $\Psi_D$, while the Video Patch Embedding and Reference Patch Embedding modules undergo fine-tuning.

During the training phase, LiftAvatar employs the Flow Matching objective consistent with Wan2.1's~\cite{DBLP:journals/corr/abs-2503-20314} training framework. The training objective is formulated as follows:

\begin{equation}
\mathcal{L} = \mathbb{E}_{x_0, x_1, c, t} \left[ \Vert u(x_t, c, t; \theta) - v_t \Vert^2 \right],
\end{equation}
where, $t$ denotes the timestep sampled from a logit-normal distribution, $x_1$ represents the latent representation of the clean driving image encoded through Wan2.1-VAE, $x_0 \sim \mathcal{N}(0, I)$ denotes random Gaussian noise, $x_t = t x_1 + (1-t) x_0$ is the intermediate latent representation, $v_t = x_1 - x_0$ is the ground-truth velocity vector, $u(x_t, c, t; \theta)$ denotes the velocity predicted by the LiftAvatar model, $c$ represents all conditional inputs, including: $I^R$, $S^R$, $S^D$, $E^R$ and $E^D$.

\begin{figure}
    \centering
    \includegraphics[width=0.95\linewidth]{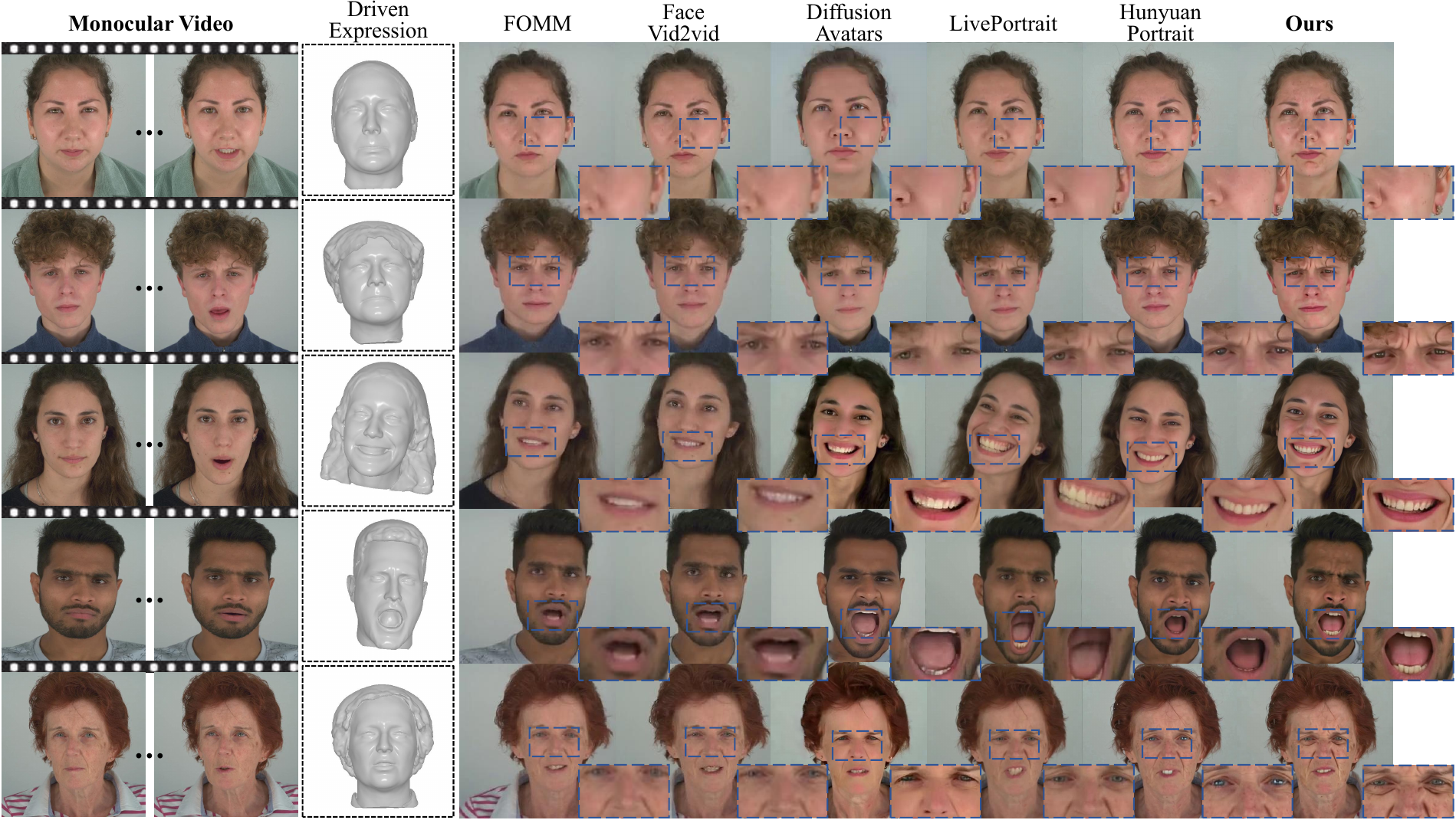}
    \caption{Qualitative results for kinematic lifted. We compared our results with non-diffusion-based methods (FOMM~\cite{DBLP:conf/nips/SiarohinLT0S19} and Face Vid2vid~\cite{DBLP:conf/cvpr/WangM021}) as well as diffusion-based models (DiffusionAvatars~\cite{DBLP:conf/cvpr/KirschsteinGN24} and LivePortrait~\cite{DBLP:journals/corr/abs-2407-03168} and HunyuanPortrait~\cite{DBLP:conf/cvpr/XuYZZJHJZCTL0L25}). It is evident that our method provides better performance in generating extreme expressions, particularly in terms of facial texture details, teeth accuracy, and pose accuracy.}
    \label{Figure_3}
\end{figure}

\begin{figure}
    \centering
    \includegraphics[width=0.95\linewidth]{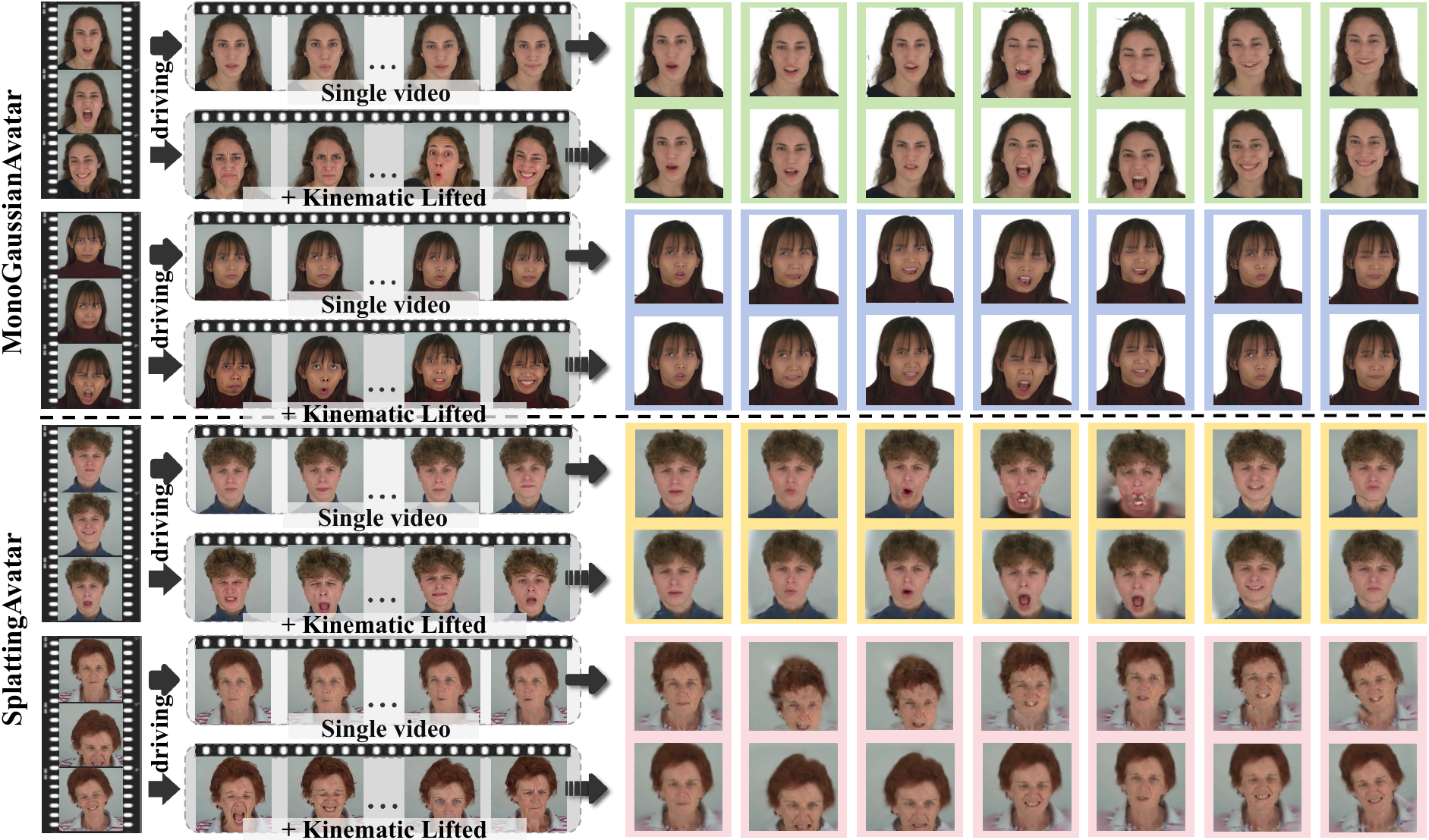}
    \caption{Qualitative results for head avatar animation. We compare the two head avatar animation methods before and after kinematic lifted. The comparison shows that our proposed strategy can effectively enhance subsequent reconstruction and driving.}
    \label{Figure_4}
\end{figure}

\begin{figure}
    \centering
    \includegraphics[width=0.95\linewidth]{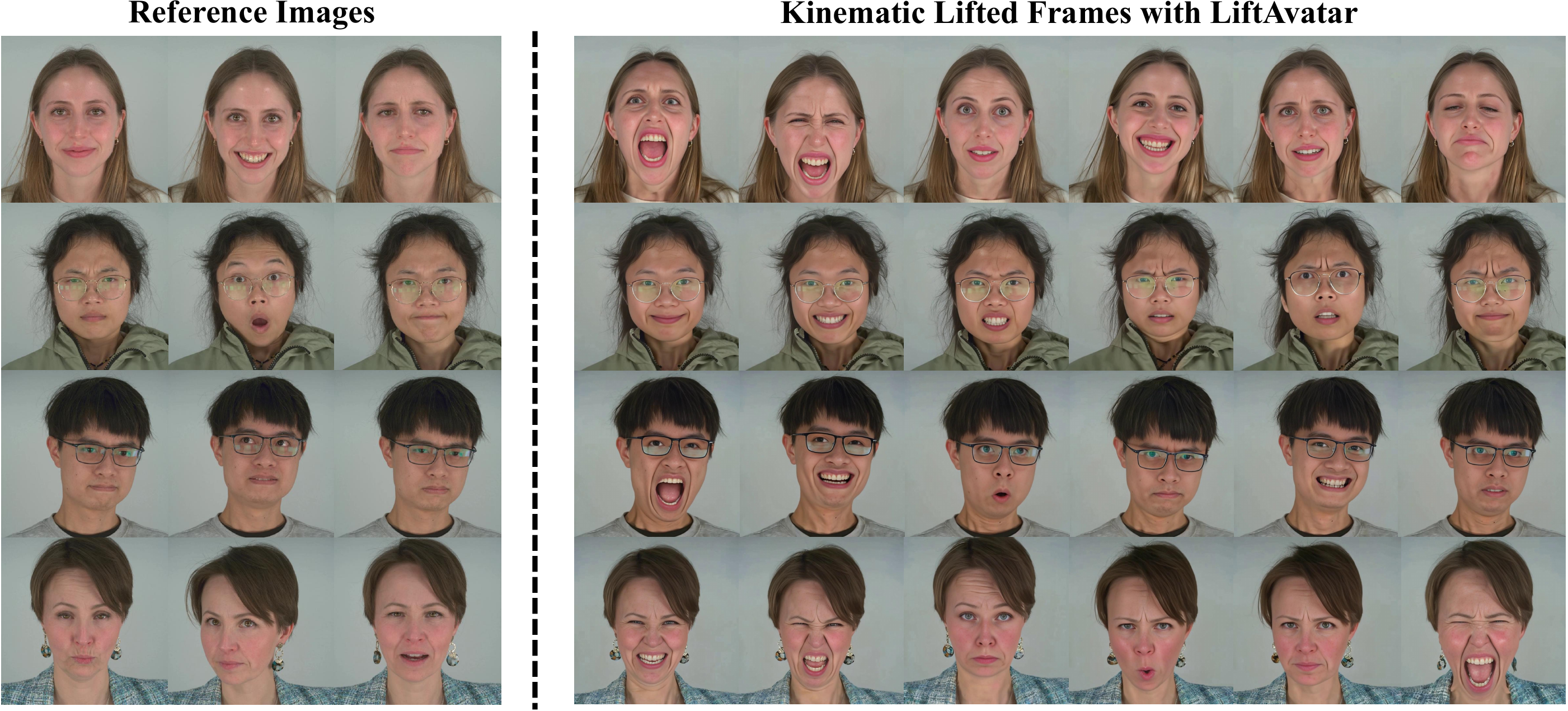}
    \caption{Lifted results with LiftAvatar.}
    \label{Figure_5}
\end{figure}

\begin{figure}
    \centering
    \includegraphics[width=0.95\linewidth]{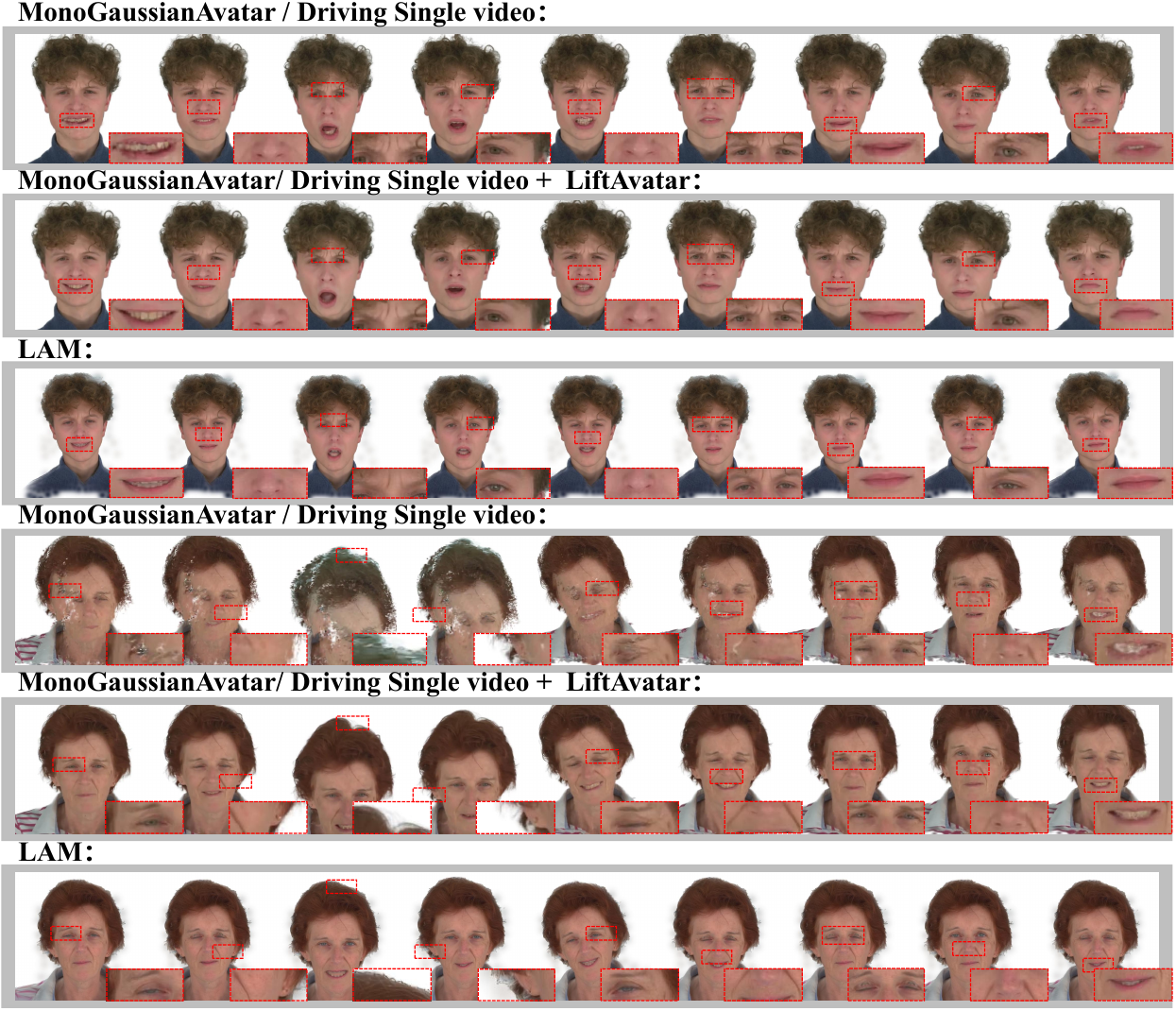}
    \caption{Additional Results from MonoGaussianAvatar and LAM.}
    \label{Figure_6}
\end{figure}

\begin{figure}
    \centering
    \includegraphics[width=0.95\linewidth]{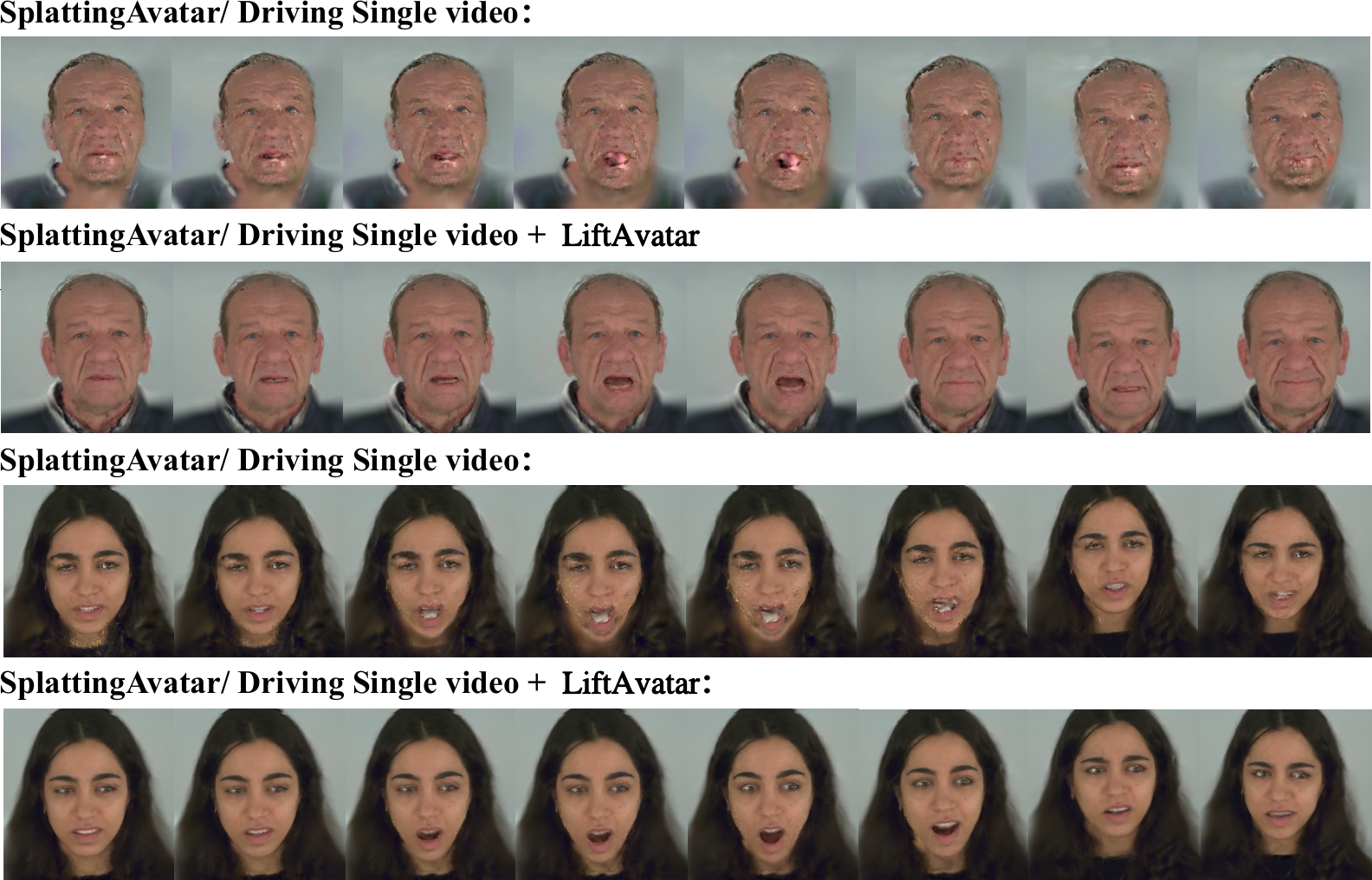}
    \caption{Additional Comparative Results of SplattingAvatar.}
    \label{Figure_7}
\end{figure}

\begin{figure}
    \centering
    \includegraphics[width=0.95\linewidth]{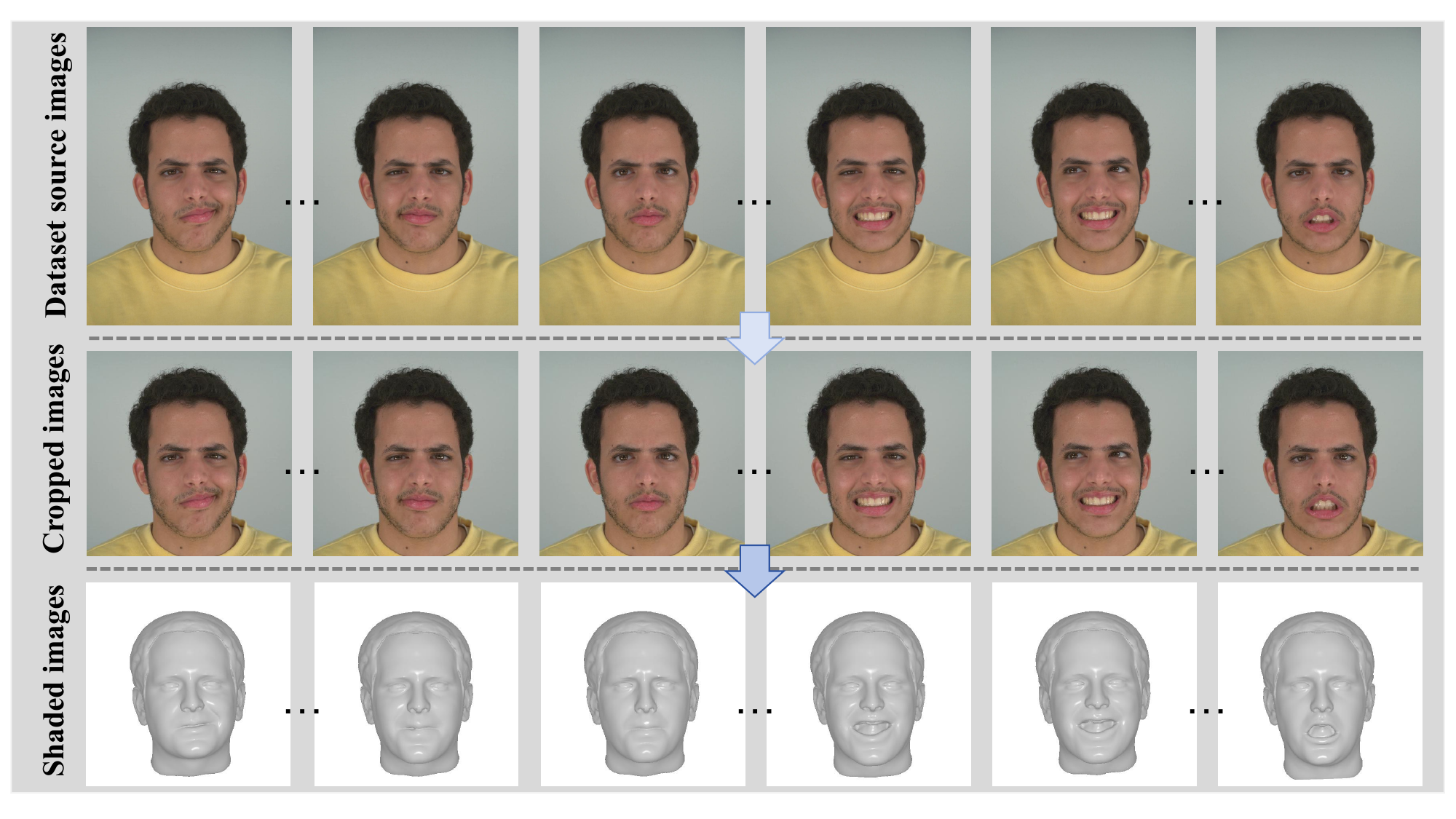}
    \caption{We first cropped the original data from $2200\times3208$ pixels to $512\times512$ pixels. Next, we performed NPHM~\cite{DBLP:conf/cvpr/GiebenhainKGRAN23} preprocessing on the cropped images, and finally, shaded them into Phong images.}
    \label{Figure_8}
\end{figure}

\section{Experiments}

\subsection{Dataset}

We conducted experiments on the NeRSemble dataset~\cite{kirschstein2023nersemble}, which contains more than 4,700 sequences involving 267 IDs, each ID corresponding to 24 sets of expression sequences. Each expression sequence includes information from 16 different angles. In total, there are 31.7 million frames of high-resolution videos covering a wide range of facial dynamics, including head movements, natural expressions, emotions, and language expressions. We first divided the different expression sequences for each ID in the dataset into training and testing sets, selecting the frontal view of all IDs as the video input. Subsequently, we fitted the NPHM~\cite{DBLP:conf/cvpr/GiebenhainKGRAN23} to the different expression sequences for these IDs. We paired different expression videos and fitted grids under predetermined angles.

\noindent\textbf{Data Crop.}
We initiate the pipeline by processing frame sequences from camera view \texttt{cam\_222200037} across all subject IDs and expression variations in the original NeRSemble dataset~\cite{kirschstein2023nersemble}. The cropping workflow consists of three core functions:
We first perform frame-by-frame cropping on the \texttt{cam\_222200037} camera angle data corresponding to all IDs in different expression sequences from the original NeRSemble dataset~\cite{kirschstein2023nersemble}. The specific functions include:
\begin{enumerate}
    \item Facial Detection and Landmark Localization \\
    Using the InsightFace library to detect faces in the images and obtain keypoints.
    
    \item Automated Cropping Region Calculation \\
    Determining a unified cropping area based on the face keypoints from multiple images.
    
    \item Image Cropping and Standardization \\
    Cropping all images according to the calculated cropping area and adjusting them to a uniform size of $512\times512$ pixels.
\end{enumerate}

\noindent\textbf{Data Generation.}
The data generation process involves reconstructing 3D face models from monocular videos and rendering them with the Phong shading model. We will illustrate the overall data processing flow for each ID, as shown in Figure~\ref{Figure_8}.
The algorithm describes the NPHM~\cite{DBLP:conf/cvpr/GiebenhainKGRAN23} monocular 3D face reconstruction and rendering pipeline, consisting of four stages:

\begin{enumerate}
    \item Preprocessing (face detection, segmentation, and landmark fitting using RetinaFace, MODNet, and PIPNet)
    \item Dynamic 3D Reconstruction (optimizing latent codes $\mathbf{z}_{\text{geo}}$ for geometry, $\mathbf{z}_{\text{app}}$ for appearance, $\mathbf{z}_{\text{exp}}^t$ for per-frame expressions, and lighting $\zeta$ via volumetric rendering)
    \item Mesh Extraction (marching cubes to extract a canonical mesh with vertices $V$ and faces $F$)
    \item Phong Rendering (deforming the mesh per frame and shading using Phong lighting model with coefficients $k_a$, $k_d$, $k_s$)
\end{enumerate}

The pipeline combines neural implicit representations with traditional mesh rendering for photorealistic results.

\subsection{Training}

\noindent\textbf{Training Setting.} 
We trained on 8$\times$NVIDIA H100 GPUs, using the AdamW optimizer. LiftAvatar is based on Wan2.1-Fun-14B Control~\cite{DBLP:journals/corr/abs-2503-20314}. LiftAvatar initializes the weights of the video patch embedding and reference patch embedding using the weights from Wan2.1's~\cite{DBLP:journals/corr/abs-2503-20314} original patch embedding, training them with a learning rate of 1e-4. For the key, query, and value embedding layers in Wan2.1's~\cite{DBLP:journals/corr/abs-2503-20314} attention module, we fine-tune using LoRA with a rank of 64, setting the learning rate to 1e-5. The batch size is configured as 2 per GPU, with training conducted over 3 days for 60,000 steps. When sampling reference images, we avoid selecting frames from the same video in the NeRSemble~\cite{kirschstein2023nersemble}. Instead, we extract frames from different videos featuring the same person to prevent the model from failing to generate expressions with significant differences.

\noindent\textbf{Kinematic Lifted Baseline.}
Our comparative analysis pits LiftAvatar against a set of non-diffusion-based (FOMM~\cite{DBLP:conf/nips/SiarohinLT0S19}, Face Vid2vid~\cite{DBLP:conf/cvpr/WangM021}) and diffusion-based (LivePortrait~\cite{DBLP:journals/corr/abs-2407-03168}, DiffusionAvatars~\cite{DBLP:conf/cvpr/KirschsteinGN24}, HunyuanPotrait~\cite{DBLP:conf/cvpr/XuYZZJHJZCTL0L25}) methods. We perform inference tests on the NeRSemble dataset~\cite{kirschstein2023nersemble}. As shown in the side-by-side visual comparison in Figure~\ref{Figure_3}, our method achieves an inference speed of 5 fps (e.g., processing 150 frames in 30 seconds). The additional time required for this inference is negligible compared to the typical several-hour training time of 3DGS avatars.

\begin{figure}
    \centering
    \includegraphics[width=0.95\linewidth]{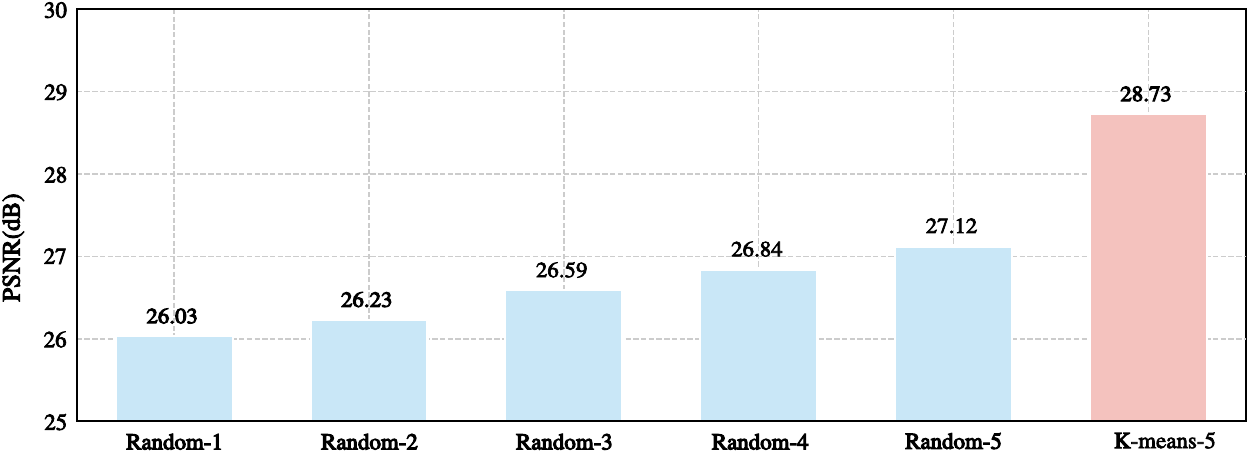} 
    \caption{The Number of Reference Images. Optimal performance is achieved with five reference frames selected by K-means (K-means-5) during inference.}
    \label{Figure_9}
\end{figure}

\begin{figure}
    \centering
    \includegraphics[width=0.95\linewidth]{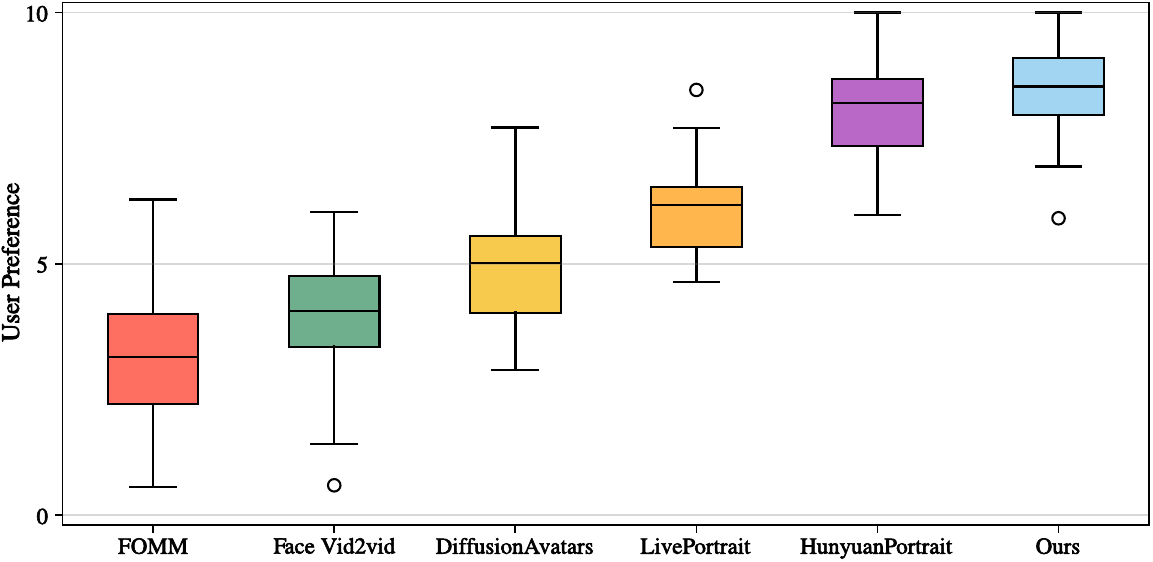} 
    \caption{User Study (50 users). In the evaluation conducted by 50 users on the kinematic lifted facial expressions, our model achieved the highest scores and significantly outperformed all other compared models.}
    \label{Figure_10}
\end{figure}

\begin{figure}
    \centering
    \includegraphics[width=0.95\linewidth]{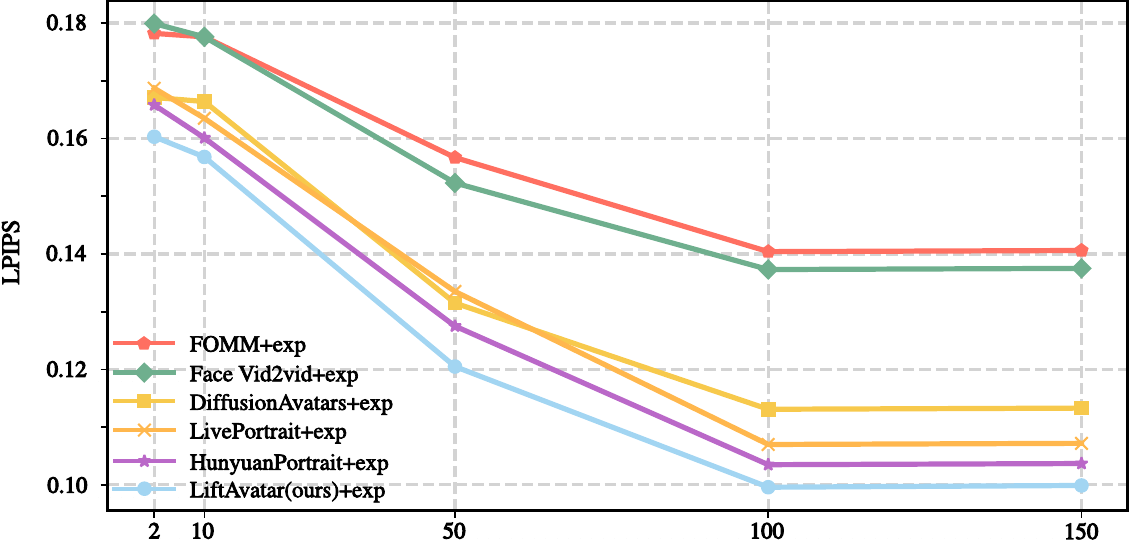}
    \caption{The Number of Kinematic Lifted Expressions. The impact of the number of expressions on the effectiveness of different kinematic lifted methods.}
    \label{Figure_11}
\end{figure}

\begin{table}[width=0.95\linewidth,cols=7,pos=h]
\caption{Quantitative comparison of expression kinematic lifted.}
\label{tab:Expression_Augmenters}
\begin{tabular*}{\tblwidth}{@{} l RRR RRR @{}}
\toprule
\multicolumn{1}{c}{\textbf{Method}} & \multicolumn{3}{c}{\textbf{Image Quality}} & \multicolumn{3}{c}{\textbf{Other Metrics}} \\
\cmidrule(lr){2-4} \cmidrule(lr){5-7}
& PSNR$\hspace{0.2em}\uparrow$ & LPIPS$\hspace{0.2em}\downarrow$ & SSIM$\hspace{0.2em}\uparrow$ & FID$\hspace{0.2em}\downarrow$ & AED$\hspace{0.2em}\downarrow$ & CSIM$\hspace{0.2em}\uparrow$ \\
\midrule
FOMM~\cite{DBLP:conf/nips/SiarohinLT0S19} & 21.9167 & 0.2047 & 0.6823 & 60.31 & 0.7433 & 0.9721 \\
Face Vid2vid~\cite{DBLP:conf/cvpr/WangM021} & 21.3089 & 0.1980 & 0.7150 & 54.61 & 0.7851 & 0.9718 \\
DiffusionAvatars~\cite{DBLP:conf/cvpr/KirschsteinGN24} & 25.3098 & 0.1479 & 0.8119 & 40.84 & 0.6525 & 0.9814 \\
LivePortrait~\cite{DBLP:journals/corr/abs-2407-03168} & 25.6561 & 0.1470 & 0.8237 & 40.55 & 0.6372 & 0.9855 \\
HunyuanPortrait~\cite{DBLP:conf/cvpr/XuYZZJHJZCTL0L25} & 27.2258 & 0.1137 & 0.8412 & 39.24 & 0.6144 & 0.9891 \\
LiftAvatar (Ours) & \textbf{28.6850} & \textbf{0.1008} & \textbf{0.8478} & \textbf{38.65} & \textbf{0.5958} & \textbf{0.9934} \\
\bottomrule
\end{tabular*}
\end{table}

\begin{table}[width=0.95\linewidth,cols=10,pos=h]
\caption{Quantitative comparisons of different expression kinematic lifted methods.}
\label{tab:Different expression augmented methods}
\begin{tabular*}{\tblwidth}{@{} l RRR RRR RRR @{}}
\toprule
\multicolumn{1}{c}{\textbf{Method}} & \multicolumn{3}{c}{FOMM+exp} & \multicolumn{3}{c}{Face Vid2vid+exp} & \multicolumn{3}{c}{DiffusionAvatars+exp} \\
\cmidrule(lr){2-4} \cmidrule(lr){5-7} \cmidrule(lr){8-10}
& PSNR$\hspace{0.2em}\uparrow$ & SSIM$\hspace{0.2em}\uparrow$ & LPIPS$\hspace{0.2em}\downarrow$ & PSNR$\uparrow$ & SSIM$\hspace{0.2em}\uparrow$ & LPIPS$\hspace{0.2em}\downarrow$ & PSNR$\hspace{0.2em}\uparrow$ & SSIM$\hspace{0.2em}\uparrow$ & LPIPS$\hspace{0.2em}\downarrow$ \\
\midrule
SplattingAvatar~\cite{DBLP:conf/cvpr/ShaoWLWL00024} & 20.92 & 0.683 & 0.247 & 21.65 & 0.710 & 0.198 & 24.96 & 0.840 & 0.145 \\
MonoGaussianAvatar~\cite{DBLP:conf/siggraph/Chen0LXZYL24} & 22.50 & 0.790 & 0.232 & 22.80 & 0.751 & 0.192 & 25.10 & 0.885 & 0.137 \\
\midrule
\multicolumn{1}{c}{\textbf{Method}} & \multicolumn{3}{c}{LivePortrait+exp} & \multicolumn{3}{c}{HunyuanPortrait+exp} & \multicolumn{3}{c}{LiftAvatar+exp} \\
\cmidrule(lr){2-4} \cmidrule(lr){5-7} \cmidrule(lr){8-10}
& PSNR$\uparrow$ & SSIM$\uparrow$ & LPIPS$\downarrow$ & PSNR$\uparrow$ & SSIM$\uparrow$ & LPIPS$\downarrow$ & PSNR$\uparrow$ & SSIM$\uparrow$ & LPIPS$\downarrow$ \\
\midrule
SplattingAvatar~\cite{DBLP:conf/cvpr/ShaoWLWL00024} & 25.06 & 0.840 & 0.186 & 25.32 & 0.843 & 0.153 & \textbf{25.71} & \textbf{0.849} & \textbf{0.132} \\
MonoGaussianAvatar~\cite{DBLP:conf/siggraph/Chen0LXZYL24} & 25.50 & 0.891 & 0.181 & 25.79 & 0.898 & 0.140 & \textbf{26.98} & \textbf{0.903} & \textbf{0.121} \\
\bottomrule
\end{tabular*}
\end{table}

\noindent\textbf{3D Head Avatar.} 
To evaluate the effectiveness of our LiftAvatar, we performed comparative experiments against state-of-the-art methods, namely SplattingAvatar~\cite{DBLP:conf/cvpr/ShaoWLWL00024} and MonoGaussianAvatar~\cite{DBLP:conf/siggraph/Chen0LXZYL24}. These methods are specifically designed for 3D head avatar reconstruction and animation, making them suitable candidates for our analysis. The experiments encompassed both basic reconstruction tasks and the rendering of novel expressions. Through a comprehensive evaluation, we are able to determine the contributions of the proposed framework in enhancing the quality of 3D head avatar representations. To ensure a fair comparison between the original input and its kinematic lifted counterpart, each group was trained for the same number of iterations under identical settings. These settings were carefully aligned with the optimal configurations provided in the official GitHub repositories. Specifically, for SplattingAvatar, we used the Adam optimizer and trained for 50,000 iterations. For MonoGaussianAvatar, we also used the Adam optimizer with a learning rate of $1 \times 10^{-4}$ and trained for 60,000 iterations, with a batch size of 16. Parameters not explicitly mentioned follow the default settings.

\subsection{Results and Comparisons}

We present the experimental results, providing a comprehensive analysis of both qualitative and quantitative outcomes. Our evaluation framework encompasses a diverse set of metrics to ensure a thorough assessment of the experimental results. The evaluation metrics we employed include Peak Signal-to-Noise Ratio (PSNR), Structural Similarity Index (SSIM), Learned Perceptual Image Patch Similarity (LPIPS), Frechet Inception Distance (FID), and Frechet Video Distance (FVD). We measure identity preservation (CSIM) by comparing the cosine similarity between the embeddings of the predicted and real images in a face recognition network. Additionally, we evaluate the performance of generative models in expression (AED) reconstruction.

\noindent\textbf{Quantitative Comparison.}
In Table~\ref{tab:Expression_Augmenters}, we compare LiftAvatar with other SOTA methods in the expression kinematic lifted task. The results are averaged over avatars generated from 10 individuals. Our method shows the best overall performance, compared to DiffusionAvatars~\cite{DBLP:conf/cvpr/KirschsteinGN24}, our method incorporates a motion module that enhances temporal consistency, eliminating the impact of artifacts, resulting in texture details that (such as hair) are closer to reality. Although LivePortrait~\cite{DBLP:journals/corr/abs-2407-03168} performs well due to the inclusion of a super-resolution module and training on a large expression dataset, it lacks control over extreme expressions and poses, often generating overly exaggerated expressions that lead to inaccuracies.
HunyuanPortrait~\cite{DBLP:conf/cvpr/XuYZZJHJZCTL0L25} also employs a large-scale video diffusion transformer, it demonstrates significant advantages over other methods. Despite these advantages, its reliance on sparse keypoints as control conditions, while offering greater flexibility, results in inferior control precision and detail preservation compared to our method.
The results indicate that images generated by LiftAvatar excel in clarity (LPIPS) and perform remarkably well on metrics such as PSNR and SSIM compared to other baseline methods. The synthesized facial features, such as teeth, eyebrows, mouth corners, and wrinkles, are more closely aligned with real expressions (AED).

\begin{table}[width=0.95\linewidth,cols=4,pos=h]
\caption{Quantitative comparisons of expression coefficient injection. +Kin-Lift means Kinematic Lifted. (w/o exp. coe.) means without expression coefficient injection, (w exp. coe.) means with expression coefficient injection.}
\label{tab:Head Avatar}
\begin{tabular*}{\tblwidth}{@{} l RRR @{}}
\toprule
\multicolumn{1}{c}{\textbf{Method}} & \multicolumn{1}{c}{PSNR$\hspace{0.2em}\uparrow$} & \multicolumn{1}{c}{SSIM$\hspace{0.2em}\uparrow$} & \multicolumn{1}{c}{LPIPS$\hspace{0.2em}\downarrow$} \\
\midrule
MonoGaussianAvatar & 19.73 & 0.689 & 0.373 \\
+Kin-Lift (w/o exp. coe.) & \underline{25.09} & \underline{0.849} & \underline{0.158} \\
+Kin-Lift (w exp. coe.) & \textbf{26.98} & \textbf{0.903} & \textbf{0.121} \\
\addlinespace[0.5em]
SplattingAvatar & 18.32 & 0.657 & 0.410 \\
+Kin-Lift (w/o exp. coe.) & \underline{24.69} & \underline{0.785} & \underline{0.264} \\
+Kin-Lift (w exp. coe.) & \textbf{25.71} & \textbf{0.849} & \textbf{0.132} \\
\bottomrule
\end{tabular*}
\end{table}

\begin{table}[width=0.95\linewidth,cols=4,pos=h]
\caption{Quantitative comparison of different reference image counts.}
\label{tab:The number of keyframes}
\begin{tabular*}{\tblwidth}{@{} l RRR @{}}
\toprule
\multicolumn{1}{c}{\textbf{Reference image}} & \multicolumn{1}{c}{PSNR$\hspace{0.2em}\uparrow$} & \multicolumn{1}{c}{FVD$\hspace{0.2em}\downarrow$} & \multicolumn{1}{c}{FID$\hspace{0.2em}\downarrow$} \\
\midrule
K-means-one & 28.59 & 341.55 & 43.87 \\
K-means-two & 28.63 & 290.94 & 42.17 \\
K-means-three & 28.65 & 263.67 & 41.69 \\
\addlinespace[0.5em]
\textbf{K-means-five} & \textbf{28.68} & \textbf{181.90} & \textbf{38.94} \\
\bottomrule
\end{tabular*}
\end{table}
In Table~\ref{tab:Different expression augmented methods}, we first apply six different expression kinematic lifted methods to the single video. We then use the kinematic lifted motion information to train MonoGaussianAvatar~\cite{DBLP:conf/siggraph/Chen0LXZYL24} and SplattingAvatar~\cite{DBLP:conf/cvpr/ShaoWLWL00024}, resulting in the corresponding trained models. Finally, we drive these models with unseen extreme expressions and compare their performance with the models trained on the original monocular video. The results indicate that our proposed kinematic lifted strategy is effective, and thanks to the expression injection in LiftAvatar, and the reference images injection, our method demonstrates superior overall quantitative metrics.

\noindent\textbf{Qualitative Comparison.}
We further provide a qualitative comparison in Figure~\ref{Figure_3}.
We apply different baseline methods for expression kinematic lifted on monocular videos of different IDs. In contrast, our LiftAvatar demonstrates superior performance in generating texture details in facial regions, including the mouth, eyebrows, and teeth. Even in areas where the NPHM~\cite{DBLP:conf/cvpr/GiebenhainKGRAN23} mesh is poorly explained, our method can reasonably fill these regions, exhibiting excellent 3D consistency, precise expression generation, and a realistic appearance.
In Figure~\ref{Figure_4}, we compare two state-of-the-art head avatar methods, and it is visually evident that the driving effects achieved through expression kinematic lifted outperform those from single video driving. This further demonstrates the effectiveness of our LiftAvatar.
We provide additional kinematic lifted results Figure~\ref{Figure_5}.It is evident that our LiftAvatar can effectively generate highly natural and high-quality extreme expression images, even when the input Reference image has a limited range of expressions.To further facilitate the reproducibility of our LiftAvatar, we will open-source all the involved code and data.
We also provide additional visual comparisons in Figure~\ref{Figure_6}, comparing the results of: (a) our MonoGaussianAvatar driven by a single video, (b) driving with facial expressions enhanced by LiftAvatar, and (c) the LAM~\cite{he2025lam} method. Although LAM~\cite{he2025lam} enables the direct generation of animatable and renderable Gaussian head models from a single image, and its feed-forward architecture aligns with current trends in generative models, it still exhibits limitations in fine-grained detail generation. Moreover, existing approaches in this line have yet to establish an efficient and reliable pathway for fully leveraging the information contained in monocular videos. Therefore, the method we propose remains of significant research value and practical potential. Figure~\ref{Figure_7} shows the comparative effect of SplattingAvatar. It is evident that the kinematic lifted training framework we proposed is effective.

\noindent\textbf{User Study.} 
We conducted a user study with 50 participants to subjectively evaluate the kinematic lifted effect of our method (LiftAvatar) against five state-of-the-art approaches: FOMM~\cite{DBLP:conf/nips/SiarohinLT0S19}, Face Vid2vid~\cite{DBLP:conf/cvpr/WangM021}, DiffusionAvatars~\cite{DBLP:conf/cvpr/KirschsteinGN24}, LivePortrait~\cite{DBLP:journals/corr/abs-2407-03168}, and HunyuanPortrait~\cite{DBLP:conf/cvpr/XuYZZJHJZCTL0L25}. Each participant rated the generated videos on a scale from 0 (worst) to 10 (best), and the results are summarized in Figure~\ref{Figure_10}. Our method achieved the highest mean score of $8.5 \pm 0.8$, substantially outperforming HunyuanPortrait~\cite{DBLP:conf/cvpr/XuYZZJHJZCTL0L25} ($8.0 \pm 1.0$), LivePortrait~\cite{DBLP:journals/corr/abs-2407-03168} ($6.0 \pm 0.9$), DiffusionAvatars~\cite{DBLP:conf/cvpr/KirschsteinGN24} ($5.0 \pm 1.1$), Face Vid2vid~\cite{DBLP:conf/cvpr/WangM021} ($4.0 \pm 1.3$), and FOMM~\cite{DBLP:conf/nips/SiarohinLT0S19} ($3.5 \pm 1.5$). Notably, our approach not only attained the highest average rating but also exhibited the smallest variance, indicating consistent user preference. A one-way analysis of variance (ANOVA) confirmed a significant effect of the method on ratings ($F(5,294) = 96.3$, $p < 0.001$), and post-hoc Tukey HSD tests revealed that our method significantly outperforms all other methods ($p < 0.001$ for each pairwise comparison). These results demonstrate that the lifted effects generated by our approach are perceived as more natural and visually appealing than those of existing techniques.


\subsection{Ablation Study}
\label{sec:experiments}

\noindent\textbf{Expression Coefficient Injection.} 
In Table~\ref{tab:Head Avatar}, we study whether injecting NPHM expression coefficients is necessary for downstream 3D head avatar reconstruction and animation. In particular, we compare training with the original single video and with the expression kinematic lifted video under two settings: \emph{(w/o exp. coe.)} removes the coefficient injection branch, while \emph{(w exp. coe.)} keeps it enabled.
The results show a clear and consistent advantage of coefficient injection. Intuitively, shading maps provide fine-grained, pixel-level deformation cues, but they are not always sufficient to disambiguate expression semantics (e.g., subtle mouth-corner movement versus cheek deformation), especially under extreme or unseen expressions. Injecting expression coefficients supplies a structured, low-dimensional and semantically aligned control signal, which stabilizes the diffusion-driven motion generation and reduces drifting or over-exaggeration. 
Consequently, both MonoGaussianAvatar and SplattingAvatar exhibit noticeably improved reconstruction quality and animation fidelity after enabling coefficient injection. Without this enhancement, the lifted training data may still contain imprecise expression trajectories, leading to suboptimal avatar deformation and perceptual artifacts. More importantly, integrating expression injection consistently improves all evaluation metrics, indicating that coefficient injection is not merely a minor add-on but a key component for high-quality and controllable head avatar animation.

\noindent\textbf{Reference Image Selection.} 
In the inference process, selecting reference images is crucial because reference frames anchor identity, appearance details (e.g., skin texture, teeth, hair), and illumination cues for the diffusion model. A single reference image is often insufficient to cover complex facial states, and may force the model to over-rely on priors, resulting in over-smoothed details or identity drift. 
To address this, we adopt an automatic and diversity-aware selection strategy. Specifically, we apply K-means clustering on the 100-dimensional expression coefficient space and select the frames closest to cluster centers as reference images. This strategy reduces redundancy among references and maximizes expression coverage within the observed video, thereby improving the conditioning quality.
As shown in Table~\ref{tab:The number of keyframes}, increasing the number of reference images generally improves inference results, as more complementary cues are provided. When using five reference images, we observe consistent improvements across all metrics compared to using only a single reference image. This gain is also reflected in Figure~\ref{Figure_9}, where randomly selected references lead to fluctuating performance, while the K-means-selected set yields more stable and stronger results. We further evaluate $k=\{2,4,6,7\}$ to identify the best trade-off. When $k<5$, the reference set lacks diversity, limiting the model's ability to faithfully reconstruct or synthesize challenging expressions. When $k>5$, the computational cost increases significantly due to the expanded conditioning context, yet the visual quality gain becomes marginal. Therefore, $k=5$ provides the optimal balance between performance and efficiency in our setting.

\noindent\textbf{Number of Kinematic Lifted Expressions.} 
We further investigate how the number of kinematic lifted expressions affects the final quality. Figure~\ref{Figure_11} shows that increasing the number of lifted expressions consistently reduces LPIPS, indicating improved perceptual similarity and better reconstruction/animation realism. This trend aligns with our motivation: more lifted expressions effectively enrich the kinematic coverage of the training data, reducing overfitting to a narrow motion manifold and improving generalization to unseen or extreme expressions.
Notably, the improvement exhibits diminishing returns. The steepest quality gain occurs when the number of expressions increases from 2 to 50, suggesting that early expansion of expression coverage rapidly alleviates kinematic sparsity. After reaching 100 expressions, the curve flattens, indicating that the downstream avatar pipeline approaches a saturation regime where additional expressions contribute less new information. This observation suggests a practical guideline: choosing a moderate number of lifted expressions can already deliver most of the benefits, while further increasing the number mainly increases training cost and storage without proportional gains in perceptual quality.

\section{Conclusion}
\label{sec:conclusion}

Existing methods for 3D avatar reconstruction primarily focus on improving the models themselves. However, in this paper, we approach the task from a completely different perspective by lifting input data within the kinematic space. Specifically, we enhance monocular videos, which often suffer from limited expression and pose variations, thereby reducing the difficulty of downstream reconstruction tasks. We propose a novel kinematic lifted training method called LiftAvatar, built upon a video generation architecture. LiftAvatar directly completes the expressions and poses of monotonous video inputs, significantly enriching the information content of the input data and integrating large-scale, high-quality data priors. Extensive experiments demonstrate that LiftAvatar substantially enhances the model's expressive capability in 3DGS~\cite{DBLP:journals/tog/KerblKLD23} when dealing with monotonous input videos. It improves the stability of extreme expressions, reduces the occurrence of artifacts, and significantly lowers the model's requirements for data quality and diversity.


\section*{Declaration of competing interest}
The authors declare that they have no known competing finacial interests or personal relationships that could have appeared to influence the work reported in this paper.

\section*{Data availability}
Data will be made available on request.

\printcredits

\bibliographystyle{cas-model2-names}

\bibliography{cas-refs}






\end{document}